\def\year{2019}\relax
\documentclass[letterpaper]{article} 
\usepackage{aaai19}  
\usepackage{times}  
\usepackage{helvet}  
\usepackage{courier}  
\usepackage{url}  
\usepackage{graphicx}  
\usepackage{tabulary}
\usepackage{booktabs}
\usepackage{amsmath}
\usepackage{CJKutf8}
\usepackage{amsfonts}
\usepackage{amssymb}
\usepackage{CJK}
\newcommand{\cev}[1]{\reflectbox{\ensuremath{\vec{\reflectbox{\ensuremath{#1}}}}}}
\newcommand{\namecite}[1]{\citeauthor{#1}~\shortcite{#1}}

\begin{document}
\begin{CJK*}{UTF8}{gbsn}
%
\title{Neural Melody Composition from Lyrics}

\author{
    Hangbo Bao$^{1}$\thanks{\;Contribution during internship at Microsoft Research.}, 
    Shaohan Huang$^2$, 
    Furu Wei$^2$, \\
    \textbf{\Large{Lei Cui$^2$,}} 
    \textbf{\Large{Yu Wu$^3$,}}
    \textbf{\Large{Chuanqi Tan$^3$}}, 
    \textbf{\Large{Songhao Piao}}$^1$, 
    \textbf{\Large{Ming Zhou}}$^2$
\\ 
$^1$School of Computer Science, Harbin Institute of Technology, Harbin, China\\
$^2$Microsoft Research, Beijing, China\\
$^3$State Key Laboratory of Software Development Environment, Beihang University, Beijing, China\\
addf400@foxmail.com,\{shaohanh,fuwei,lecu,mingzhou\}@microsoft.com, \\
wuyu@buaa.edu.cn, tanchuanqi@nlsde.buaa.edu.cn, piaosh@hit.edu.cn
}

\maketitle
\begin{abstract}

In this paper, we study a novel task that learns to compose music from natural language. Given the lyrics as input, we propose a melody composition model that generates lyrics-conditional melody as well as the exact alignment between the generated melody and the given lyrics simultaneously. More specifically, we develop the melody composition model based on the sequence-to-sequence framework. It consists of two neural encoders to encode the current lyrics and the context melody respectively, and a hierarchical decoder to jointly produce musical notes and the corresponding alignment. 
Experimental results on lyrics-melody pairs of 18,451 pop songs demonstrate the effectiveness of our proposed methods. In addition, we apply a singing voice synthesizer software to synthesize the ``singing" of the lyrics and melodies for human evaluation. Results indicate that our generated melodies are more melodious and tuneful compared with the baseline method.

\end{abstract}

\section{Introduction}

\begin{figure}
  \begin{center}
    \includegraphics[scale=0.66]{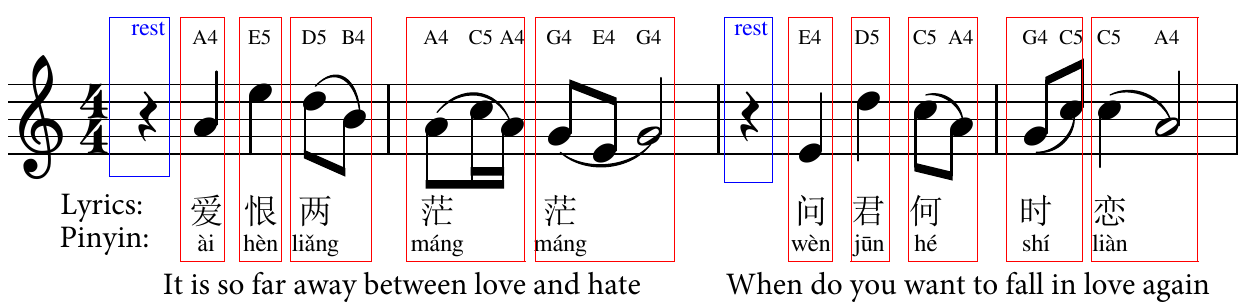}
    \caption{A fragment of a Chinese song ``Drunken Concubine (new version)''. 
    The blue rectangles indicate rests, some intervals of silence in a piece of melody. 
    The red rectangles indicate the alignment between the lyrics and the melody, meaning a mapping from syllable of lyrics to musical notes.
    Pinyin indicates the syllables for each Chinese character. 
    We can observe that the second Chinese character `恨'~(hate) aligns one note `E5' and the last Chinese character `恋'~(love) aligns two notes `C5' and `A4' in melody, 
    which describes the ``one-to-many'' relationship in the alignment between the lyrics and melody.
    }
    \label{data-example}
  \end{center}
\end{figure}

We study the task of melody composition from lyrics, which consumes a piece of text as input and aims to compose the corresponding melody as well as the exact alignment between generated melody and the given lyrics. 
Specifically, the output consists of two sequences of musical notes and lyric syllables\footnote{A syllable is a word or part of a word which contains a single vowel sound and that is pronounced as a unit. Chinese is a monosyllabic language which means words (Chinese characters) predominantly consist of a single syllable (\url{ https://en.wikipedia.org/wiki/Monosyllabic_language}).}
with two constraints. First, each syllable in the lyrics at least corresponds to one musical note in the melody. Second, a syllable in the lyrics may correspond to a sequence of notes, which increases the difficulty of this task. 
Figure~\ref{data-example} shows a fragment of a Chinese song. 
For instance, the last Chinese character `恋'~(love) aligns two notes `C5' and `A4' in the melody. 

There are several existing research works on generating lyrics-conditional melody~\cite{ackerman2017algorithmic,scirea2015smug,monteith2012automatic,fukayama2010automatic}. These works usually treat the melody composition task as a classification or sequence labeling problem. They first determine the number of musical notes by counting the syllables in the lyrics, and then predict the musical notes one after another by considering previously generated notes and corresponding lyrics.
However, 
these works only consider the ``one-to-one'' alignment between the melody and lyrics. 
According to our statistics on 18,451 Chinese songs, $97.9\%$ songs contains at least one syllable that corresponds to multiple musical notes (i.e. ``one-to-many'' alignment), thus the simplification may introduce bias into the task of melody composition. 

In this paper, we propose a novel melody composition model which can generate melody from lyrics and well handle the ``one-to-many'' alignment between the generated melody and the given lyrics. 
For the given lyrics as input, we first divide the input lyrics into sentences and then use our model to compose each piece of melody from the sentences one by one. Finally, we merge these pieces to a complete melody for the given lyrics. 
More specifically, it consists of two encoders and one hierarchical decoder. The first encoder encodes the syllables in current lyrics into an array of hidden vectors with a bi-directional recurrent neural network (RNN) and the second encoder leverages an attention mechanism to convert the context melody into a dynamic context vector with a two-layer bi-directional RNN. In the decoder, we employ a three-layer RNN decoder to produce the musical notes and the alignment jointly, where the first two layers are to generate the pitch and duration of each musical note and the last layer is to predict a label
for each generated musical note to indicate the alignment. 

We collect 18,451 Chinese pop songs and generate the lyrics-melody pairs with precise syllable-note alignment to conduct experiments on our methods and baselines. 
Automatic evaluation results show that our model outperforms baseline methods on all the metrics. In addition, we leverage a singing voice synthesizer software to synthesize the “singing” of the lyrics and melodies and ask human annotators to manually judge the quality of the generated pop songs.
The human evaluation results further indicate that the generated lyrics-conditional melodies from our method are more melodious and tuneful compared with the baseline methods. 

The contributions of our work in this paper are summarized as follows.

  \begin{itemize}
    \item To the best of our knowledge, this paper is the first work to use end-to-end neural network model to compose melody from lyrics.
    \item We construct a large-scale lyrics-melody dataset with 18,451 Chinese pop songs and 644,472 lyrics-context-melody triples, so that the neural networks based approaches are possible for this task.
    \item Compared with traditional sequence-to-sequence models, our proposed method can generate the exact alignment as well as the ``one-to-many'' alignment between the melody and lyrics. 
    \item The human evaluation verifies that the synthesized pop songs of the generated melody and input lyrics are melodious and meaningful.
  \end{itemize}

\section{Preliminary}
We first introduce some basic definitions from music theory and then give a brief introduction to our lyrics-melody parallel corpus. Table~\ref{Notations} lists some mathematical notations used in this paper. 

\subsection{Concepts from Music Theory}
Melody can be regarded as an ordered sequence of many musical notes. The basic unit of melody is the musical note which mainly consists of two attributes: pitch and duration. The pitch is a perceptual property of sounds that allows their ordering on a frequency-related scale, or more commonly, the pitch is the quality that makes it possible to judge sounds as ``higher'' and ``lower'' in the sense associated with musical melodies\footnote{\url{https://en.wikipedia.org/wiki/Pitch_(music)}}. 
Therefore, we use a sequence of numbers to represent the pitch. For example, we represent `C5' and `Eb6' as 72 and 87 respectively based on the MIDI\footnote{\url{https://newt.phys.unsw.edu.au/jw/notes.html}}. 
A rest is an interval of silence in a piece of music and we use `$R$' to represent it and treat it as a special pitch. 
Duration is a particular time interval to describe the length of time that the pitch or tone sounds\footnote{\url{https://en.wikipedia.org/wiki/Duration_(music)}}, which is to judge how long or short a musical note lasts.

\subsection{Lyrics-Melody Parallel Corpus}

\begin{figure}
    \begin{center}
      \begin{flushleft}
          An example of a sheet music:
      \end{flushleft}
      \includegraphics[scale=0.66]{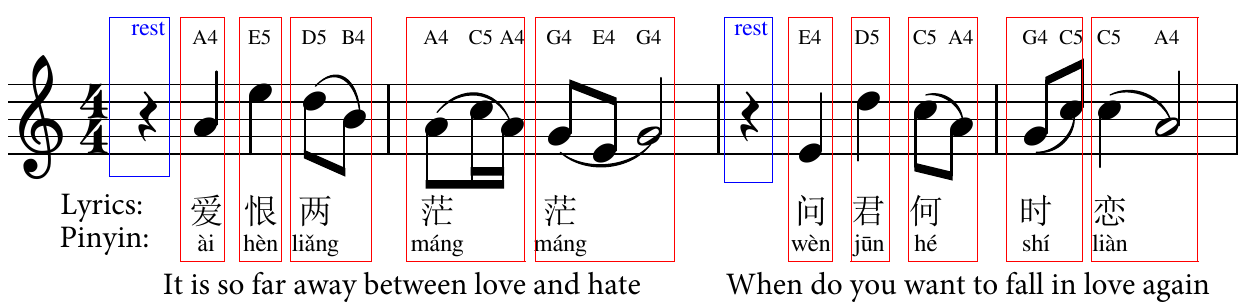}
      \begin{flushleft}
        Lyrics-melody aligned data:
      \end{flushleft}

        \scalebox{0.55}{
            \begin{tabular}{lcccccccccccccccccccc}
                \toprule[1.5pt]
                Pinyin & 
                \multicolumn{1}{c}{ài} &
                \multicolumn{1}{c}{hèn} &
                \multicolumn{1}{c}{liǎng} &
                \multicolumn{1}{c}{máng} &
                \multicolumn{1}{c}{máng} & 
                \multicolumn{1}{c}{wèn} & 
                \multicolumn{1}{c}{jūn} & 
                \multicolumn{1}{c}{hé} & 
                \multicolumn{1}{c}{shí} & 
                \multicolumn{1}{c}{liàn} \\
                Lyrics & 
                \multicolumn{1}{c}{爱} &
                \multicolumn{1}{c}{恨} &
                \multicolumn{1}{c}{两} &
                \multicolumn{1}{c}{茫} &
                \multicolumn{1}{c}{茫} & 
                \multicolumn{1}{c}{问} & 
                \multicolumn{1}{c}{君} & 
                \multicolumn{1}{c}{何} & 
                \multicolumn{1}{c}{时} & 
                \multicolumn{1}{c}{恋} \\
                \cmidrule(lr){2-2}
                \cmidrule(lr){3-3}
                \cmidrule(lr){4-4}
                \cmidrule(lr){5-5}
                \cmidrule(lr){6-6}
                \cmidrule(lr){7-7}
                \cmidrule(lr){8-8}
                \cmidrule(lr){9-9}
                \cmidrule(lr){10-10}
                \cmidrule(lr){11-11}
                    
                $Pitch$ & R A4 & E5 & D5 B4 & A4 C5 A4 & G4 E4 G4 & 
                R E4 & D5 & C5 A4 & G4 C5 & C5 A4 \\
                $Duration$ & $\frac{1}{4}$\,\, $\frac{1}{4}$ & $\frac{1}{4}$ & $\frac{1}{8}$\,\, $\frac{1}{8}$ & $\frac{1}{8}$ $\frac{1}{16}$ $\frac{1}{16}$ & $\frac{1}{8}$\,\, $\frac{1}{8}$\,\, $\frac{1}{2}$ & 
                $\frac{1}{4}$\,\, $\frac{1}{4}$ & $\frac{1}{4}$ & $\frac{1}{8}$\,\, $\frac{1}{8}$ & $\frac{1}{8}$\,\, $\frac{1}{8}$ & $\frac{1}{4}$\,\, $\frac{1}{2}$ \\
                $Label$&0\,\,\, 1&1&0\,\,\, 1&0\,\,\, 0\,\,\,\, 1&0\,\,\, 0\,\,\, 1&0\,\,\, 1&1&0\,\,\, 1&0\,\,\, 1&0\,\,\, 1 \\
                \bottomrule[1.5pt]
            \end{tabular}
        }
        \label{tab:addlabel}
      
      \caption{An illustration for lyrics-melody aligned data. The $Pitch$ and the $Duration$ respectively represent the pitch and duration of each musical note. In addition, the $Label$ provides the information on alignment between the lyrics and melody. To be specific, a musical note is assigned with label $1$ that denotes it is a boundary of the musical note sub-sequence aligned to the corresponding syllable otherwise it is assigned with label $0$. Additionally, we always align the rests with their latter syllables.
      }
      \label{data-representation}
    \end{center}
  \end{figure}

Figure~\ref{data-representation} shows an example of a lyrics-melody aligned pair with precise syllable-note alignment, where each Chinese character of the lyrics aligns with one or more notes in the melody. 
  \begin{table}
    \caption{
      Notations used in this paper
    }
    \makebox[\linewidth][c]{
      \scalebox{0.75}{
      \begin{tabular}{ll}
        \toprule[1.5pt]
        Notations & Description \\ 
        \hline 
        $X$ & the sequence of syllables in given lyrics \\
        $x^j$ & the $j$-th syllable in $X$ \\
        $M$ & the sequence of musical notes in context melody \\
        $m^i$ & the $i$-th musical note in $M$\\
        $m_{pit}^i, m_{dur}^i$ & the pitch and duration of $m^i$, respectively\\
        $Y$ & the sequence of musical notes in predicted melody \\
        $y^i$ & the $i$-th musical note in $Y$\\
        $y_{<i}$ & the previously predicted musical notes $\{y^1, ..., y^{i-1} \}$ in $Y$\\
        $y_{pit}^i, y_{dur}^i, y_{lab}^i$ & the pitch, duration and label of $y^i$, respectively\\
        $Pitch$ & the pitch sequence comprised of each $y_{pit}^i$ in $Y$ \\
        $Duration$ & the duration sequence comprised of each $y_{dur}^i$ in $Y$ \\
        $Label$ & the label sequence comprised of each $y_{lab}^i$ in $Y$ \\
        $h_{lrc}^j$ & the $j$-th hidden state in output of lyrics encoder \\
        $h_{con}^i$ & the $i$-th hidden state in output of context melody encoder \\
        $c^i$ & the dynamic context vector at time step $i$\\
        $c_{con}^i$ & the $i$-th melody context vector from context melody encoder\\
        $R$ & indicates the rest, specially \\
        \bottomrule[1.5pt]
        \end{tabular}
      }
    }
    \label{Notations}
  \end{table}

  \begin{figure*}
    \begin{center}
      \includegraphics[width=17.8cm]{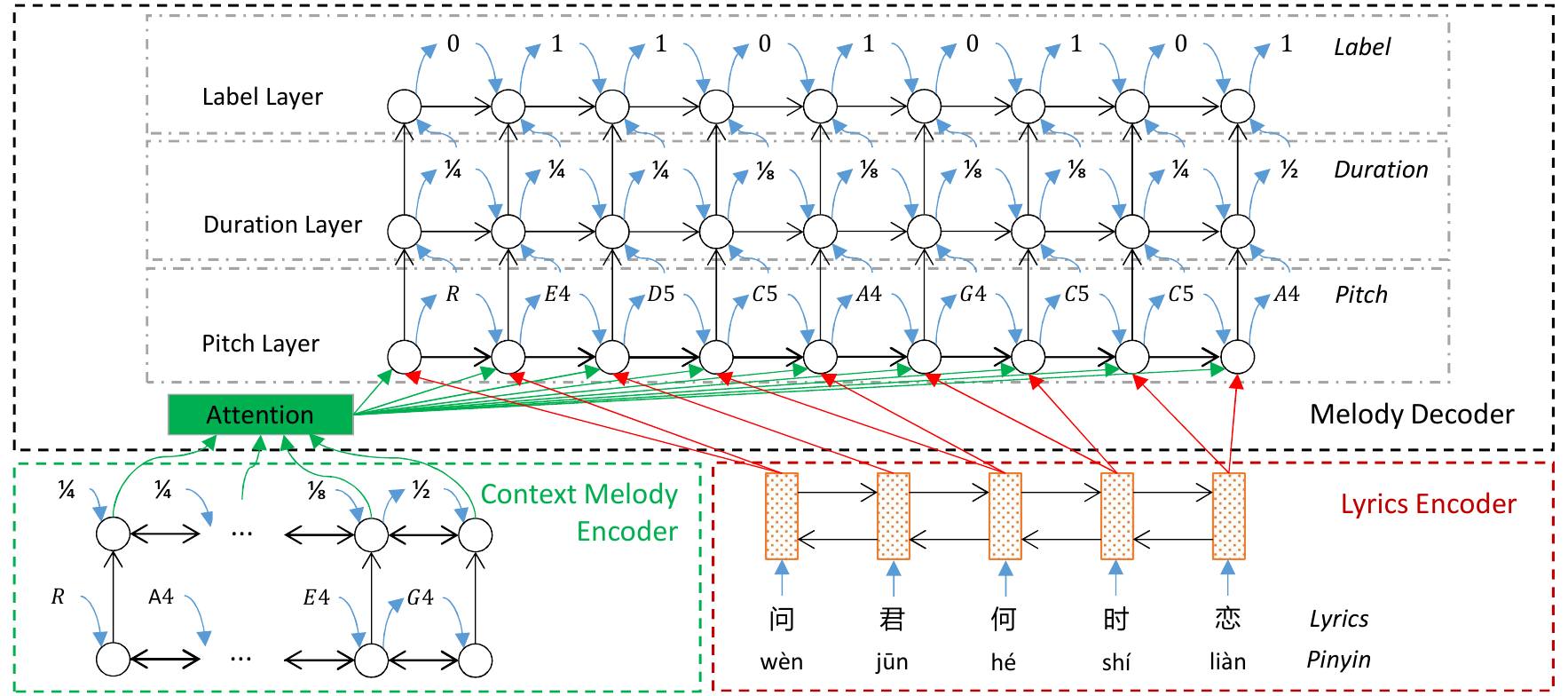}
        \caption{An illustration of Songwriter. 
        The lyrics encoder and context melody encoder encode the syllables of given lyrics and the context melody into two arrays of hidden vectors, respectively. For decoding the $i$-th musical note $y^i$, Songwriter uses attention mechanism to obtain a context vector $c_{con}^i$ from the context melody encoder~(green arrows) and counts how many label $1$ has been produced in previously musical notes to obtain $h_{con}^j$ to represent the current syllable corresponding to $y^i$ from the lyrics encoder~(red arrows) to melody decoder. In melody decoder, the pitch layer and duration layer first predict the pitch $y_{pit}^i$ and duration $y_{dur}^i$ of $y^i$, then the label layer predicts a label $y_{lab}^i$ for $y^i$ to indicate the alignment. 
        }
      \label{model-1}
    \end{center}
  \end{figure*}

The generated melody consists of three sequences $Pitch$, $Duration$ and $Label$ where the $Label$ sequence represents the alignment between melody and lyrics. We are able to rebuild the sheet music with them. $Pitch$ sequence represents the pitch of each musical note in melody and `$R$' represents the rest in $Pitch$ sequence specifically. Similarly, $Duration$ sequence represents the duration of each musical note in melody. $Pitch$ and $Duration$ consist of a complete melody but do not include information on the alignment between the given lyrics and corresponding melody. 

$Label$ contains the information of alignment. Each item of the $Label$ is labeled as one of $\{ 0, 1 \}$ to indicate the alignment between the musical note and the corresponding syllable in the lyrics. 
To be specific, a musical note is assigned with label $1$ that denotes it is a boundary of the musical note sub-sequence, which aligned to the corresponding syllable, otherwise it is assigned with label $0$. 
We can split the musical notes into the $n$ parts by label~$1$, where $n$ is the number of syllables of the lyrics, and each part is a musical note sub-sequence. 
Then we can align the musical notes to their corresponding syllables sequentially. Additionally, we always align the rests to their latter syllables. For instance, we can observe that the second rest aligns to the Chinese character `问'~(ask). 

\section{Task Definition}

Given lyrics as the input, our task is to generate the melody and alignment that make up a song with the lyrics. We can formally define this task as below:

The input is a sequence $X=(x^1, ...,x^{|X|})$ representing the syllables of lyrics. The output is a sequence $Y=(y^1, ..., y^{|Y|})$ representing the predicted musical notes for corresponding lyrics, where the $y^i = \{y_{pit}^i, y_{dur}^i, y_{lab}^i\}$. In addition, the output sequence $Y$ should satisfy the following restriction:
\begin{equation}
    |X| = \sum_{i=1}^{|Y|}y_{pit}^{i}
    \label{restriction}
\end{equation}
which restricts the generated melody can be exactly aligned with the given lyrics. 

\section{Approach}

In this section, we present the end-to-end neural networks model, termed as \textbf{Songwriter}, to compose a melody which aligns exactly to the given input lyrics. 
Figure~\ref{model-1} provides an illustration of Songwriter. 

\subsection{Overview}

Given lyrics as the input, we first divide the lyrics into sentences and then use Songwriter to compose each piece of the melody sentence by sentence. For each sentence in lyrics, Songwriter takes the syllables in the sentence lyrics and the context melody, which are some previous predicted musical notes, as input and then predicts a piece of melody. When the last piece of melody has been predicted, we merge these pieces of melody to make a complete song with the given lyrics. This procedure can be considered as a sequence generation problem with two sequences as input, syllables of the current lyrics $X$ and the context melody $M$. We develop our melody composition model based on a modified RNN encoder-decoder~\cite{cho-EtAl:2014:EMNLP2014} to support multiple sequences as input.

Songwriter employs two neural encoders, lyrics encoder and context melody encoder, to respectively encode the syllables of the current lyrics $X$ and the context melody $M$, and leverages a hierarchical melody decoder to produce musical notes and the alignment $Y$. 
To be specific, the lyrics encoder and context melody encoder encode $X$ and $M$ into two arrays of hidden vectors, respectively. 
At the time step $i$, melody decoder obtains a context vector $c_{con}^i$ from the context melody encoder and a hidden vector $h_{lrc}^j$ from the lyrics encoder to produce the $i$-th musical note $y^i$. $c_{con}^i$ is computed dynamically by the attention mechanism from the output of the context melody encoder. 
$h_{lrc}^j$ is one of output hidden vectors of the lyrics encoder, which represents the $j$-th syllable $x^j$ in the current lyrics. 
In the melody decoder, which is a three-layer RNN, the pitch layer and duration layer first predict the pitch $y_{pit}^i$ and duration $y_{dur}^i$, then the label layer predicts a label $y_{lab}^i$ of $y^i$ to indicate the alignment. 

\subsection{Gated Recurrent Units}

We use Gated Recurrent Unit (GRU)~\cite{ChoMGBBSB14} instead of basic RNN. We describe the mathematical model of the GRU as follows: 

\begin{eqnarray}
  &&z^i = \sigma(\mathbf{W_{hz}}h^{i-1} + \mathbf{W_{xz}}x^i + \mathbf{b_z}) \\
  &&r^i = \sigma(\mathbf{W_{hr}}h^{i-1} + \mathbf{W_{xr}}x^i + \mathbf{b_r}) \\
  &&\widehat{h}^i = \mathrm{tanh}\big(\mathbf{W_{h}}(r^i\circ h^{i-1}) + \mathbf{W_x} x^i + \mathbf{b}\big) \\
  &&h^i = (1-z^i)\circ h^{i-1} + z^i\circ \widehat{h}^i
\end{eqnarray}
where $\mathbf{W_{hz}}$, $\mathbf{W_{xz}}$, $\mathbf{b_z}$, $\mathbf{W_{hr}}$, $\mathbf{W_{xr}}$, $\mathbf{b_r}$, $\mathbf{W_{h}}$, $\mathbf{W_x}$ and $\mathbf{b}$ are parameters to be learned in GRU, $\circ$ is an element-wise multiplication, $\sigma(\cdot)$ is a logistic sigmoid function, $r^i$ and $z^i$ are the gates and $h^i$ is the hidden state at time step $i$.

\subsection{Lyrics Encoder}

We use a bi-directional RNN~ \cite{Schuster:1997:BRN:2198065.2205129} built by two GRUs to encode the syllables of lyrics which concatenates the syllable feature embedding and word embedding as input $X = \{x^1, ..., x^{|X|}\}$ to the GRU encoders: 
\begin{eqnarray}
&& \vec{h}_{lrc}^i = f_\mathrm{GRU}(\vec{h}_{lrc}^{i-1}, x^i) \\
&& \cev{h}_{lrc}^i = f_\mathrm{GRU}(\cev{h}_{lrc}^{i+1}, x^i) \\
&& h_{lrc}^i = \left[ \begin{array}{c}
\vec{h}_{lrc}^{i} \\
\cev{h}_{lrc}^{i}
\end{array}
\right]
\end{eqnarray}
Then, the lyrics encoder outputs an array of hidden vectors $\{h_{lrc}^1, ..., h_{lrc}^{|X|}\}$ to represent the information of each syllable in the lyrics. 
\subsection{Context Melody Encoder}
  
We use the context melody encoder to encode the context melody $M = \{m^1, ..., m^{|M|}\}$. The encoder is a two-layer RNN that encodes pitch, duration and label of a musical note respectively at each time step. Each layer is a bi-directional RNN which is built by two GRUs. For the first layer, we describe the forward directional GRU and the backward directional GRU at time step $i$ as follows: 
  \begin{eqnarray}
    && \vec{h}_{pit}^i = f_\mathrm{GRU}(\vec{h}_{pit}^{i-1}, m_{pit}^i) \\
    && \cev{h}_{pit}^i = f_\mathrm{GRU}(\cev{h}_{pit}^{i+1}, m_{pit}^i)
  \end{eqnarray}
where $m_{pit}^i$ is the pitch attribute of $i$-th note $m^i$. Then, we concatenate them into one vector: 
  \begin{equation}
    h_{pit}^i = \left[ \begin{array}{c}
      \vec{h}_{pit}^{i}\\
      \cev{h}_{pit}^{i}
    \end{array}
    \right]
  \end{equation}
The bottom layer encodes the output of the first layer and the duration attribute of melody. The employment can be described as follows: 
\begin{eqnarray}
&&  \vec{h}_{dur}^i = f_\mathrm{GRU}(\vec{h}_{dur}^{i-1}, m_{dur}^i, h_{pit}^{i}) \\
&&  \cev{h}_{dur}^i = f_\mathrm{GRU}(\cev{h}_{dur}^{i+1}, m_{dur}^i, h_{pit}^{i}) \\
&&  h_{dur}^i = \left[ \begin{array}{c}
  \vec{h}_{dur}^{i}\\
  \cev{h}_{dur}^{i}
    \end{array}
\right]
\end{eqnarray}
We concatenate the two output arrays of vectors to an array of vectors to represent the context melody sequence: 
  \begin{equation}
    h_{con}^i = \left[ \begin{array}{c}
      h_{pit}^{i}\\
      h_{dur}^{i}
    \end{array}
    \right]
  \end{equation}
  
\subsection{Melody Decoder}
  
The decoder predicts the next note $y^{i}$ from all previously predicted notes $\{y^1, ..., y^{i-1} \}$ ($y_{<i}$, for short), the context musical notes $M = \{m^1, ..., m^{|M|} \}$ and the syllables $X = \{x^1, ..., x^{|X|}\}$ of given lyrics. We define the conditional probability when decoding $i$-th note as follows: 
\begin{equation}
    arg~max~P(y^i|y_{<i}, X, M)
    \label{conditional-probability-all}
\end{equation}
  To model the three attributes of $y^{i}$, where we use $\{ y_{pit}^{i}, y_{dur}^{i}, y_{lab}^{i}\}$ to respectively represent the pitch, duration and label, we decompose Eq. (\ref{conditional-probability-all}) into Eq.(\ref{conditional-probability-decomposed}): 
  \begin{equation}
      \label{conditional-probability-decomposed}
    \begin{aligned}
      P(y^i|y_{<i}, X, M) = &P(y_{pit}^i|y_{<i}, X, M)~\cdot \\
      &P(y_{dur}^i|y_{<i}, X, M, y_{pit}^i)~\cdot \\
      &P(y_{lab}^i|y_{<i}, X, M, y_{pit}^i, y_{dur}^i)
    \end{aligned}
  \end{equation}
We use a three-layer RNN as decoder to respectively decode the pitch, duration and label of a musical note at each time step. We define the conditional probabilities of each layer in the decoder: 
\begin{equation}
    P(y_{pit}^i|y_{<i}, X, M) = g_p(s_{pit}^i, c^i, y^{i-1})
\end{equation}
\begin{equation}
    P(y_{dur}^i|y_{<i}, X, M, y_{pit}^i) = g_d(s_{dur}^i, c^i, y^{i-1}, y_{pit}^i)
\end{equation}
\begin{equation}
P(y_{lab}^i|y_{<i}, X, M, y_{pit}^i, y_{dur}^i) = g_l(s_{lab}^i, c^i, y^{i-1}, y_{pit}^i, y_{dur}^i)
\end{equation}
where $g_p(\cdot)$, $g_d(\cdot)$ and $g_l(\cdot)$ are nonlinear functions that output the probabilities of $y_{pit}^i$, $y_{dur}^i$ and $y_{lab}^i$ respectively. $s_{pit}^i$, $s_{dur}^i$ and $s_{lab}^i$ are respectively the corresponding hidden states of each layer. $c^i$ is a dynamic context vector representing the $M$ and $X$. We introduce the employment of $c^i$ before $s_{pit}^i$, $s_{dur}^i$ and $s_{lab}^i$: 

\begin{equation}
    c^i = c_{con}^i + h_{lrc}^j
\end{equation}
where $c_{con}^i$ is a context vector from context melody encoder and $h_{lrc}^j$ is one of output hidden vectors of lyrics encoder, which represent the $x_j$ that should be aligned to the current predicting $y^i$. In particular, we set $c_{con}^i$ as a zero vector if there is no context melody as input.
From our representation method for lyrics-melody aligned pairs, it is not difficult to understand how to get the $x^j$ that $y^i$ should be aligned to: 
\begin{equation}
  \label{Label-Based-Precise-Aligning}
  j = \sum_{t=1}^{i-1}y_{lab}^t
\end{equation}
$c_{con}^i$ is recomputed at each step by alignment model~\cite{DBLP:journals/corr/BahdanauCB14} as follows:
\begin{equation}
  c_{con}^{i} = \sum_{t = 1}^{|M|}\alpha^{i,t}h_{con}^{t}
\end{equation}
where $h_{con}^{t}$ is one hidden vector from the output of melody encoder and the weight $\alpha_{i,t}$ is computed by:
\begin{eqnarray}
  &&  \alpha^{i,t} = \frac{exp(e^{i,t})}{\sum_{k = 1}^{|M|}exp(e^{i,k})} \\
  &&  e_{i,k} = \mathbf{v_{a}}^\intercal tanh(\mathbf{W_{a}}s^{i-1} + \mathbf{U_{a}}h_{con}^{k})
\end{eqnarray}
where $\mathbf{v_{a}}$, $\mathbf{W_{a}}$ and $\mathbf{U_{a}}$ are learnable parameters. Finally, we obtain the $c^i$ and then employ the $s_p^i$, $s_d^i$, $s_l^i$ and $s^i$ as follows: 
\begin{eqnarray}
s_{pit}^{i} = f_\mathrm{GRU}(s_{pit}^{i - 1}, c^{i - 1}, y_{pit}^{i - 1}, h_{lrc}^{j}) \\
s_{dur}^{i} = f_\mathrm{GRU}(s_{dur}^{i - 1}, c^{i - 1}, y_{dur}^{i - 1}, y_{pit}^i, s_{pit}^{i}) \\
s_{lab}^{i} = f_\mathrm{GRU}(s_{lab}^{i - 1}, c^{i - 1}, y_{lab}^{i - 1}, y_{pit}^i, y_{dur}^i, d_{dur}^{i}) \\
s^i = 
[
    {s_p^{i}}^\intercal;
    {s_d^{i}}^\intercal;
    {s_l^{i}}^\intercal
]^\intercal 
\end{eqnarray}
  
\subsection{Objective Function}
Given a training dataset with $n$ lyrics-context-melody triples $\mathcal{D} = \{X^{(i)}, M^{(i)}, Y^{(i)}\}^{n}_{i=1}$, where $X^{(i)} = \{x^{(i)1}, ..., x^{(i)|X^{(i)}|}\}$, $M^{(i)} = \{m^{(i)1}, ..., m^{(i)|M^{(i)}|}\}$ 
and $Y^{(i)} = \{y^{(i)1}, ..., y^{(i)|Y_{(i)}|}\}$. In addition, $\forall (i,j)$, $y^{(i)j} = (y_{pit}^{(i)j}, y_{dur}^{(i)j}, y_{lab}^{(i)j})$. 
Our training objective is to minimize the negative log likelihood loss $\mathcal{L}$ with respect to the learnable model parameter $\theta$: 
\begin{equation}
  \mathcal{L} = - \frac{1}{n}\sum_{i=1}^{n}\sum_{j=1}^{|Y^{(i)}|}log~P(y_{pit}^{(i)j},y_{dur}^{(i)j},y_{lab}^{(i)j}|\theta,X^{(i)},M^{(i)},y_{<j})
\end{equation}
where $y_{<j}$ is short for $\{y_{(i)}^1, ..., y_{(i)}^{j}\}$.
  
  \begin{center}
    \begin{table*}[h]
    \caption{
      Automatic evaluation results
    }
    \makebox[\linewidth][c]{
      \scalebox{1}{
        \begin{tabular}{lcccccccccccccc}
          \toprule[1.5pt]
          & \multicolumn{10}{c}{Teacher-forcing} & \multicolumn{2}{c}{Sampling}\\
          \cmidrule(lr){2-11}
          & & \multicolumn{3}{c}{Pitch} & \multicolumn{3}{c}{Duration} & \multicolumn{3}{c}{Label} \\
          \cmidrule(lr){3-5} \cmidrule(lr){6-8} \cmidrule(lr){9-11} \cmidrule(lr){12-13}
          & $\mathrm{PPL}$ & $\mathrm{P}$ & $\mathrm{R}$ & $\mathrm{F_1}$ & $\mathrm{P}$ & $\mathrm{R}$ & $\mathrm{F_1}$ & $\mathrm{P}$ & $\mathrm{R}$ & $\mathrm{F_1}$ & $\mathrm{BLEU}$ & $\mathrm{DW}$ \\
          \hline
          CRF & / & 41.23 & 42.02 & 40.98 & 49.82 & 53.12 & 50.84 & / & / & / & 2.02 & 25.53\\ 
          Seq2seq & 2.21 & 54.76 & 55.01 & 54.56 & 64.66 & 67.88 & 65.33 & 93.14 & 93.06 & 92.60 & 3.96 & 37.04 \\
          \textbf{Songwriter} & \textbf{2.01} & \textbf{63.23} & \textbf{63.24} & \textbf{62.90} & \textbf{69.18} & \textbf{71.28} & \textbf{69.69} & \textbf{93.54} & \textbf{93.61} & \textbf{93.31} & \textbf{6.63} & \textbf{38.31} \\ 
          \bottomrule[1.5pt]
        \end{tabular}
      }
    }
    \label{Automatic}
  \end{table*}
  \end{center}
  
\section{Experiments}
  
\subsection{Dataset}

We crawled 18,451 Chinese pop songs, which include melodies with the duration over 800 hours in total, from an online Karaoke app. Then preprocess the dataset with rules as described in~\namecite{zhu2018xiaoice} to guarantee the reliability of the melodies. For each song, we convert the melody to C major or A minor that can keep all melodies in the same tune and we set BPM (Beats Per Minute) to 60 to calculate the duration of each musical note in the melody. We further divide the lyrics into sentences with their corresponding musical notes as lyrics-melody pairs. Besides, we set a window size as 40 to the context melody and use the previously musical notes as the context melody for each lyrics-melody pair to make up lyrics-context-melody triples. Finally, we obtain 644,472 triples to conduct our experiments. 
We randomly choose $5\%$ songs for validating, $5\%$ songs for testing and the rest of them for training. 

\subsection{Baselines}

As melody composition task can generally be regarded as a sequence labeling problem or a machine translation problem, we select two state-of-the-art models as baselines. 

\begin{itemize}
    \item{\textbf{CRF}} A modified sequence labeling model based on CRF~\cite{lafferty2001conditional} which contains two layers for predicting $Pitch$ and $Duration$, respectively. For ``one-to-many'' relationships, this model uses some special tags to represent a series of original tags. For instance, if a syllable aligns two notes `C5' and 'A4', we use a tag `C5A4' to represent them.
    \item{\textbf{Seq2seq}} A modified attention based sequence to sequence model which contains two encoders and one decoder. Compared with Songwriter, Seq2seq uses attention mechanism~\cite{DBLP:journals/corr/BahdanauCB14} to capture information on the given lyrics. 
    Seq2seq may not guarantee the alignment between the generated melody and syllables in given lyrics. To avoid this problem, Seq2seq model stops predicting when the number of the label $1$ in predicted musical notes is equal to the number of syllables in the given lyrics. 
  \end{itemize}

\subsection{Implementation}
\subsubsection{Model Size}
For all the models used in this paper, the number of recurrent hidden units is set to 256. In the context melody encoder and melody decoder, we treat the $pitch$, $duration$, and $label$ as tokens and use word embedding to represent them with 128, 128, and 64 dimensions, respectively. In the lyrics encoder, we use GloVe~\cite{pennington2014glove} to pre-train a char-level word embedding with 256 dimensions on a large Chinese lyrics corpus and use Pinyin\footnote{\url{https://en.wikipedia.org/wiki/Pinyin}} as the syllable features with 128 dimensions. 
\subsubsection{Parameter Initialization}
We use two linear layers with the last backward hidden states of the context melody encoder to respectively initialize the hidden states of the pitch layer and duration layer in the melody decoder in Songwriter and Seq2seq. We use zero vectors to initialize the hidden states in the lyrics encoder and context melody encoder. 

\subsubsection{Training}
We use Adam~\cite{diederik2015adam} with an initial learning rate of $0.001$ and an exponential decay rate of $0.9999$ as the optimizer to train our models with batch size as $64$, and we use the cross entropy as the loss function. 

\subsection{Automatic evaluation}
We use two modes to evaluate our model and baselines. 
\begin{itemize}
  \item \textbf{Teacher-forcing}: As in~\cite{roberts2018hierarchical}, models use the ground truth as input for predicting the next-step at each time step.
  \item \textbf{Sampling} Models predict the melody from given lyrics without any ground truth.
\end{itemize}

\subsubsection{Metrics} We use the $F_1$ score to the automatic evaluation from \namecite{roberts2018hierarchical}. Additionally, we select three automatic metrics for our evaluation as follows. 

\begin{itemize}
    \item \textbf{Perplexity~(PPL)} This metric is a standard evaluation measure for language models and can measure how well a probability model predicts samples. Lower PPL score is better.
    \item \textbf{(weighted)}~\textbf{Precision}, \textbf{Recall} and $\mathbf{F_1}$\footnote{We calculate these metrics by scikit-learn with the parameter average set as `weighted’:~\url{http://scikit-learn.org/stable/modules/classes.html#module-sklearn.metrics}} These metrics measure the performance of predicting the different attributes of the musical notes.
    \item \textbf{BLEU} This metric~\cite{papineni2002bleu} is widely used in machine translation. We use it to evaluate our predicted pitch. Higher BLEU score is better.

    \item \textbf{Duration of Word (DW)} This metric checks the sum of the duration of all notes which aligned to one word is equal to the ground truth. Higher DW score is better. 
\end{itemize}

\subsubsection{Results}

The results of the automatic evaluation are shown in Table \ref{Automatic}. We can see that our proposed method outperforms all models in all metrics. As Songwriter performs better than Seq2seq, it shows that the exact information of the syllables~(Eq.~(\ref{Label-Based-Precise-Aligning})) can enhance the quality of predicting the corresponding musical notes relative to attention mechanism in traditional Seq2seq models. In addition, the CRF model demostrates lower performance in all metrics. In CRF model, we use a special tag to represent multiple musical notes if a syllable aligns more than one musical note, which will produce a large number of different kinds of tags and result in the CRF model is difficult to learn from the sparse data. 

\subsection{Human evaluation}
Similar to the text generation and dialog response generation~\cite{zhang2014chinese,schatzmann2005quantitative}, it is challenging to accurately evaluate the quality of music composition results with automatic metrics. To this end,
we invite 3 participants as human annotators to evaluate the generated melodies from our models and the ground truth melodies of human creations.
We randomly select $20$ lyrics-melody pairs, the average duration of each melody approximately 30 seconds, from our testing set. For each selected pair, we prepare three melodies, ground truth of human creations and the generated results from Songwriter and Seq2seq. Then, we synthesized all melodies with the lyrics by NiaoNiao \footnote{A singing voice synthesizer software which can synthesize Chinese song, \url{http://www.dsoundsoft.com/product/niaoeditor/}} using default settings for the generated songs and ground truth, which is to eliminate the influences of other factors of singing. As a result, we obtain 3 (annotators) $\times$ 3 (melodies) $\times$ 20 (lyrics) samples in total. The human annotations are conducted in a blind-review mode, which means that human annotators do not know the source of the melodies during the experiments.
\subsubsection{Metrics}
We use the metrics from previous work on human evaluation for music composition as shown below. 
We also include an \textit{emotion} score to measure the relationship between the generated melody and the given lyrics. 
The human annotators are asked to rate a score from 1 to 5 after listening to the songs. Larger scores indicate better quality in all the three metrics. 

\begin{itemize}
  \item \textbf{Emotion} Does the melody represent the emotion of the lyrics?
  \item \textbf{Rhythm}~\cite{zhu2018xiaoice,watanabe2018melody} When listening to the melody, are the duration and pause of words natural?
  \item \textbf{Overall}~\cite{watanabe2018melody} What is the overall score of the melody?
\end{itemize}

\begin{table}
  \caption{
    Human evaluation results in blind-review mode
  }
  \makebox[\linewidth][c]{
    \scalebox{1}{
    \begin{tabular}{llll}
      \toprule
      Model & Overall & Emotion & Rhythm \\
      \hline 
      Seq2seq & 3.28 & 3.52 & 2.66 \\ 
      \textbf{Songwriter} & \textbf{3.83} & \textbf{3.98} & \textbf{3.52} \\ 
      Human &  4.57 & 4.50 & 4.17 \\ 
      \bottomrule 
      \end{tabular}
    }
  }
  \label{table-evaluation}
\end{table}

\subsubsection{Results}
Table~\ref{table-evaluation} shows the human evaluation results. According to the results, Songwriter outperforms Seq2seq in all metrics, which indicates its effectiveness over the Seq2seq baseline.
On the ``Rhythm'' metrics, human annotators give significantly lower scores to Seq2seq than Songwriter, which shows that the generated melodies from Songwriter are more natural on the pause and duration of words than the ones generated by Seq2seq. The results further suggest that using the exact information of syllables (Eq.~(\ref{Label-Based-Precise-Aligning})) is more effective than the soft attention mechanism in traditional Seq2seq models in the melody composition task. 
We can also observe from Table~\ref{table-evaluation} that the gaps between the system generated melodies and the ones created by human are still large on all the three metrics. It remains an open challenge for future research to develop better algorithms and models to generate melodies with higher quality. 

\section{Related Work}

A variety of music composition works have been done over the last decades. Most of the traditional methods compose music based on music theory and expert domain knowledge. 
\namecite{chan2006improving} design rules from music theory to use music clips to stitch them together in a reasonable way. With the development of machine learning and the increase of public music data, data-driven methods such as Markov chains model ~\cite{pachet2011markov} and graphic model ~\cite{pachet2017sampling} have been introduced to compose music.
  
Recently, deep learning has been revealed the potentials for musical creation. Most of these deep learning approaches use the recurrent neural network (RNN) to compose the music by regarding as a sequence. The MelodyRNN~\cite{waite2016generating} model, proposed by Google Brain Team, uses looking back RNN and attention RNN to capture the long-term dependency of melody. 
\namecite{chu2016song}
propose a hierarchical RNN based model which additionally incorporates knowledge from music theory into the representation to compose not only the melody but also the drums and chords. 
Some recent works have also started exploring various generative adversarial networks (GAN) models to compose music ~\cite{mogren2016c,yang2017midinet,dong2017musegan}. \citeauthor{brunner2018midi} \shortcite{brunner2018midi} design recurrent variational autoencoders (VAEs) with a hierarchical decoder to reproduce short musical sequences. 

Generating a lyrics-conditional melody is a subset of music composition but under more restrictions. 
Early works first determine the number of musical notes by counting the syllables in lyrics and then predict the musical notes one after another by considering previously generated notes and corresponding lyrics. 
\citeauthor{fukayama2010automatic} \shortcite{fukayama2010automatic} use dynamic programming to compute a melody from Japanese lyrics, the calculation needs three human well-designed constraints. 
\citeauthor{monteith2012automatic} \shortcite{monteith2012automatic} propose a melody composition pipeline for given lyrics. For each given lyrics, it first generates hundreds of different possibilities for rhythms and pitches. Then it ranks these possibilities with a number of different metrics in order to select a final output.
\citeauthor{scirea2015smug} \shortcite{scirea2015smug} employ Hidden Markov Models (HMM) to generate rhythm based on the phonetics of the lyrics already written. Then a harmonical structure is generated, followed by generation of a melody matching the underlying harmony. 
\citeauthor{ackerman2017algorithmic} \shortcite{ackerman2017algorithmic} design a co-creative automatic songwriting system ALYSIA base on machine learning model using random forests, which analyzes the lyrics features to generate one note at a time for each syllable. 
  
\section{Conclusion and Future Work}

In this paper, we propose a lyrics-conditional melody composition model which can generate melody and the exact alignment between the generated melody and the given lyrics. 
We develop the melody composition model under the encoder-decoder framework, which consists of two RNN encoders, lyrics encoder and context melody encoder, and a hierarchical RNN decoder. The lyrics encoder encodes the syllables of current lyrics into a sequence of hidden vectors. The context melody leverages an attention mechanism to encode the context melody into a dynamic context vector. In the decoder, it uses two layers to produce musical notes and another layer to produce alignment jointly. 
Experimental results on our dataset, which contains 18,451 Chinese pop songs, demonstrate our model outperforms baseline models. Furthermore, we leverage a singing voice synthesizer software to synthesize “singing” of the lyrics and generated melodies for human evaluation. Results indicate that our generated melodies are more melodious and tuneful. 
For future work, we plan to incorporate the emotion and the style of lyrics to compose the melody. 
      
      \small
      \bibliographystyle{aaai}
      \bibliography{acl2017}

\begin{thebibliography}{}

\bibitem[\protect\citeauthoryear{Ackerman and
  Loker}{2017}]{ackerman2017algorithmic}
Ackerman, M., and Loker, D.
\newblock 2017.
\newblock Algorithmic songwriting with alysia.
\newblock In {\em International Conference on Evolutionary and Biologically
  Inspired Music and Art},  1--16.
\newblock Springer.

\bibitem[\protect\citeauthoryear{Bahdanau, Cho, and
  Bengio}{2014}]{DBLP:journals/corr/BahdanauCB14}
Bahdanau, D.; Cho, K.; and Bengio, Y.
\newblock 2014.
\newblock Neural machine translation by jointly learning to align and
  translate.
\newblock {\em CoRR} abs/1409.0473.

\bibitem[\protect\citeauthoryear{Brunner \bgroup et al\mbox.\egroup
  }{2018}]{brunner2018midi}
Brunner, G.; Konrad, A.; Wang, Y.; and Wattenhofer, R.
\newblock 2018.
\newblock Midi-vae: Modeling dynamics and instrumentation of music with
  applications to style transfer.
\newblock In {\em Proc. Int. Society for Music Information Retrieval Conf}.

\bibitem[\protect\citeauthoryear{Chan, Potter, and
  Schubert}{2006}]{chan2006improving}
Chan, M.; Potter, J.; and Schubert, E.
\newblock 2006.
\newblock Improving algorithmic music composition with machine learning.
\newblock In {\em Proceedings of the 9th International Conference on Music
  Perception and Cognition, ICMPC}.

\bibitem[\protect\citeauthoryear{Cho \bgroup et al\mbox.\egroup
  }{2014a}]{cho-EtAl:2014:EMNLP2014}
Cho, K.; van Merrienboer, B.; Gulcehre, C.; Bahdanau, D.; Bougares, F.;
  Schwenk, H.; and Bengio, Y.
\newblock 2014a.
\newblock Learning phrase representations using rnn encoder--decoder for
  statistical machine translation.
\newblock In {\em Proceedings of the 2014 Conference on Empirical Methods in
  Natural Language Processing (EMNLP)},  1724--1734.
\newblock Doha, Qatar: Association for Computational Linguistics.

\bibitem[\protect\citeauthoryear{Cho \bgroup et al\mbox.\egroup
  }{2014b}]{ChoMGBBSB14}
Cho, K.; van Merrienboer, B.; G{\"{u}}l{\c{c}}ehre, {\c{C}}.; Bahdanau, D.;
  Bougares, F.; Schwenk, H.; and Bengio, Y.
\newblock 2014b.
\newblock Learning phrase representations using {RNN} encoder-decoder for
  statistical machine translation.
\newblock In {\em Proceedings of the 2014 Conference on Empirical Methods in
  Natural Language Processing, {EMNLP} 2014, October 25-29, 2014, Doha, Qatar,
  {A} meeting of SIGDAT, a Special Interest Group of the {ACL}},  1724--1734.

\bibitem[\protect\citeauthoryear{Chu, Urtasun, and Fidler}{2016}]{chu2016song}
Chu, H.; Urtasun, R.; and Fidler, S.
\newblock 2016.
\newblock Song from pi: A musically plausible network for pop music generation.
\newblock {\em arXiv preprint arXiv:1611.03477}.

\bibitem[\protect\citeauthoryear{Diederik P.~Kingma}{2015}]{diederik2015adam}
Diederik P.~Kingma, J.~B.
\newblock 2015.
\newblock Adam: A method for stochastic optimization.
\newblock {\em In Proceedings of the International Conference on Learning
  Representations (ICLR)}.

\bibitem[\protect\citeauthoryear{Dong \bgroup et al\mbox.\egroup
  }{2017}]{dong2017musegan}
Dong, H.-W.; Hsiao, W.-Y.; Yang, L.-C.; and Yang, Y.-H.
\newblock 2017.
\newblock Musegan: Symbolic-domain music generation and accompaniment with
  multi-track sequential generative adversarial networks.
\newblock {\em arXiv preprint arXiv:1709.06298}.

\bibitem[\protect\citeauthoryear{Fukayama \bgroup et al\mbox.\egroup
  }{2010}]{fukayama2010automatic}
Fukayama, S.; Nakatsuma, K.; Sako, S.; Nishimoto, T.; and Sagayama, S.
\newblock 2010.
\newblock Automatic song composition from the lyrics exploiting prosody of the
  japanese language.
\newblock In {\em Proc. 7th Sound and Music Computing Conference (SMC)},
  299--302.

\bibitem[\protect\citeauthoryear{Lafferty, McCallum, and
  Pereira}{2001}]{lafferty2001conditional}
Lafferty, J.; McCallum, A.; and Pereira, F.~C.
\newblock 2001.
\newblock Conditional random fields: Probabilistic models for segmenting and
  labeling sequence data.

\bibitem[\protect\citeauthoryear{Mogren}{2016}]{mogren2016c}
Mogren, O.
\newblock 2016.
\newblock C-rnn-gan: Continuous recurrent neural networks with adversarial
  training.
\newblock {\em arXiv preprint arXiv:1611.09904}.

\bibitem[\protect\citeauthoryear{Monteith, Martinez, and
  Ventura}{2012}]{monteith2012automatic}
Monteith, K.; Martinez, T.~R.; and Ventura, D.
\newblock 2012.
\newblock Automatic generation of melodic accompaniments for lyrics.
\newblock In {\em ICCC},  87--94.

\bibitem[\protect\citeauthoryear{Pachet and Roy}{2011}]{pachet2011markov}
Pachet, F., and Roy, P.
\newblock 2011.
\newblock Markov constraints: steerable generation of markov sequences.
\newblock {\em Constraints} 16(2):148--172.

\bibitem[\protect\citeauthoryear{Pachet, Papadopoulos, and
  Roy}{2017}]{pachet2017sampling}
Pachet, F.; Papadopoulos, A.; and Roy, P.
\newblock 2017.
\newblock Sampling variations of sequences for structured music generation.
\newblock In {\em Proceedings of the 18th International Society for Music
  Information Retrieval Conference (ISMIR’2017), Suzhou, China},  167--173.

\bibitem[\protect\citeauthoryear{Papineni \bgroup et al\mbox.\egroup
  }{2002}]{papineni2002bleu}
Papineni, K.; Roukos, S.; Ward, T.; and Zhu, W.-J.
\newblock 2002.
\newblock Bleu: a method for automatic evaluation of machine translation.
\newblock In {\em Proceedings of the 40th annual meeting on association for
  computational linguistics},  311--318.
\newblock Association for Computational Linguistics.

\bibitem[\protect\citeauthoryear{Pennington, Socher, and
  Manning}{2014}]{pennington2014glove}
Pennington, J.; Socher, R.; and Manning, C.
\newblock 2014.
\newblock Glove: Global vectors for word representation.
\newblock In {\em Proceedings of the 2014 conference on empirical methods in
  natural language processing (EMNLP)},  1532--1543.

\bibitem[\protect\citeauthoryear{Roberts \bgroup et al\mbox.\egroup
  }{2018}]{roberts2018hierarchical}
Roberts, A.; Engel, J.; Raffel, C.; Hawthorne, C.; and Eck, D.
\newblock 2018.
\newblock A hierarchical latent vector model for learning long-term structure
  in music.
\newblock {\em arXiv preprint arXiv:1803.05428}.

\bibitem[\protect\citeauthoryear{Schatzmann, Georgila, and
  Young}{2005}]{schatzmann2005quantitative}
Schatzmann, J.; Georgila, K.; and Young, S.
\newblock 2005.
\newblock Quantitative evaluation of user simulation techniques for spoken
  dialogue systems.
\newblock In {\em 6th SIGdial Workshop on DISCOURSE and DIALOGUE}.

\bibitem[\protect\citeauthoryear{Schuster and
  Paliwal}{1997}]{Schuster:1997:BRN:2198065.2205129}
Schuster, M., and Paliwal, K.
\newblock 1997.
\newblock Bidirectional recurrent neural networks.
\newblock {\em Trans. Sig. Proc.} 45(11):2673--2681.

\bibitem[\protect\citeauthoryear{Scirea \bgroup et al\mbox.\egroup
  }{2015}]{scirea2015smug}
Scirea, M.; Barros, G.~A.; Shaker, N.; and Togelius, J.
\newblock 2015.
\newblock Smug: Scientific music generator.
\newblock In {\em ICCC},  204--211.

\bibitem[\protect\citeauthoryear{Waite}{2016}]{waite2016generating}
Waite, E.
\newblock 2016.
\newblock Generating long-term structure in songs and stories.
\newblock {\em Magenta Bolg}.

\bibitem[\protect\citeauthoryear{Watanabe \bgroup et al\mbox.\egroup
  }{2018}]{watanabe2018melody}
Watanabe, K.; Matsubayashi, Y.; Fukayama, S.; Goto, M.; Inui, K.; and Nakano,
  T.
\newblock 2018.
\newblock A melody-conditioned lyrics language model.
\newblock In {\em Proceedings of the 2018 Conference of the North American
  Chapter of the Association for Computational Linguistics: Human Language
  Technologies, Volume 1 (Long Papers)}, volume~1,  163--172.

\bibitem[\protect\citeauthoryear{Yang, Chou, and Yang}{2017}]{yang2017midinet}
Yang, L.-C.; Chou, S.-Y.; and Yang, Y.-H.
\newblock 2017.
\newblock Midinet: A convolutional generative adversarial network for
  symbolic-domain music generation.
\newblock {\em arXiv preprint arXiv:1703.10847}.

\bibitem[\protect\citeauthoryear{Zhang and Lapata}{2014}]{zhang2014chinese}
Zhang, X., and Lapata, M.
\newblock 2014.
\newblock Chinese poetry generation with recurrent neural networks.
\newblock In {\em EMNLP},  670--680.

\bibitem[\protect\citeauthoryear{Zhu \bgroup et al\mbox.\egroup
  }{2018}]{zhu2018xiaoice}
Zhu, H.; Liu, Q.; Yuan, N.~J.; Qin, C.; Li, J.; Zhang, K.; Zhou, G.; Wei, F.;
  Xu, Y.; and Chen, E.
\newblock 2018.
\newblock Xiaoice band: A melody and arrangement generation framework for pop
  music.
\newblock In {\em Proceedings of the 24th ACM SIGKDD International Conference
  on Knowledge Discovery \& Data Mining},  2837--2846.
\newblock ACM.

\end{thebibliography}
    \end{CJK*}
\end{document}